\documentclass[lettersize,journal]{IEEEtran}

\usepackage{booktabs}

\usepackage{amsmath,amsfonts}
\usepackage{algorithmic}
\usepackage{algorithm}
\usepackage{array}
\usepackage[caption=false,font=normalsize,labelfont=sf,textfont=sf]{subfig}
\usepackage{textcomp}
\usepackage{stfloats}
\usepackage{url}
\usepackage{verbatim}
\usepackage{graphicx}
\usepackage{cite}
\begin{document}

\title{
	CapsuleBot: A Novel Hybrid Aerial-Ground Bi-Copter Robot with Two Actuated-Wheel-Rotors
	}

\author{Zhi Zheng, Qifeng Cai, Jin Wang, Xinhang Xu, Muqing Cao, Huan Yu, Jihao Li, Jun Meng and Guodong Lu%
	\thanks{Manuscript received: July, 10, 2024; Revised September, 8, 2024; Accepted November, 10, 2024. This paper was recommended for publication by Editor C. Gosselin upon evaluation of the Associate Editor and Reviewers' comments. This work was supported in part by "Pioneer" and "Leading Goose" R\&D Program of Zhejiang under Grant 2024C01170, the National Natural Science Foundation of China under Grant 52475033, and Robotics Institute of Zhejiang University under Grant K12107 and K11805. \textit{(Zhi Zheng and Qifeng Cai contributed equally to this work.) (Corresponding author: Jin Wang.)}}
	\thanks{
		Zhi Zheng, Qifeng Cai, Jin Wang, Huan Yu, Jihao Li and Guodong Lu are with The State Key Laboratory of Fluid Power and Mechatronic Systems, School of Mechanical Engineering, Zhejiang University, Hangzhou 310027, China, also with Robotics Institute of Zhejiang University, Zhejiang University, Hangzhou 310027, China, and also with  Robotics Research Center of Yuyao City, Ningbo 315400, China(e-mail: z.z@zju.edu.cn; 22225022@zju.edu.cn; dwjcom@zju.edu.cn; h.yu@zju.edu.cn; 12225048@zju.edu.cn; lugd@zju.edu.cn).}
	\thanks{
		Xinhang Xu is with School of Electrical and Electronic Engineering, Nanyang Technological University, 50 Nanyang Avenue, Singapore 639798, Singapore(e-mail: xu0021ng@e.ntu.edu.sg).}
	\thanks{	
		Muqing Cao is with Robotics Institute, Carnegie Mellon University, 5000 Forbes Avenue, Pittsburgh 15217, United States(e-mail: caom0006@e.ntu.edu.sg).}
	\thanks{
		Jun Meng is with Center for Data Mining and Systems Biology, College of Electrical Engineering, Zhejiang University, Hangzhou 310027, China.}
	\thanks{Digital Object	Identifier (DOI): see top of this page.}
}

\markboth{IEEE Robotics and Automation Letters. Preprint Version. Accepted November, 2024}
{Zheng \MakeLowercase{\textit{et al.}}: CapsuleBot: A Novel Hybrid Aerial-Ground Bi-Copter Robot with Two Actuated-Wheel-Rotors} 
\maketitle

\begin{abstract}
This paper presents the design, modeling, and experimental validation of CapsuleBot, a novel hybrid aerial-ground bi-copter robot designed for long-endurance and low-noise operations. CapsuleBot combines the maneuverability of a bi-copter in the air with the low power consumption and low noise of a two-wheel self-balancing robot on the ground. To achieve this, we design an innovative mechanical structure named the actuated-wheel-rotor, which uses a servo motor and a brushless motor to function as both a tilting rotor in the air and an actuated wheel on the ground. CapsuleBot is equipped with two actuated-wheel-rotors, enabling it to achieve hybrid aerial-ground propulsion using only four motors, with no additional motors required compared to a bi-copter. Additionally, we develop comprehensive dynamics and control systems for both air and wheel mode, based on the bi-copter model and the two-wheel self-balancing robot model. A prototype of CapsuleBot is constructed, and its performance in terms of low power consumption and low noise is validated through experiments. Challenging tasks demonstrate CapsuleBot's capability to climb steep, fly over cliffs, and traverse rough terrains.
\end{abstract}


\begin{figure}[t]
	\begin{center}
		\includegraphics[width=1.0\columnwidth]{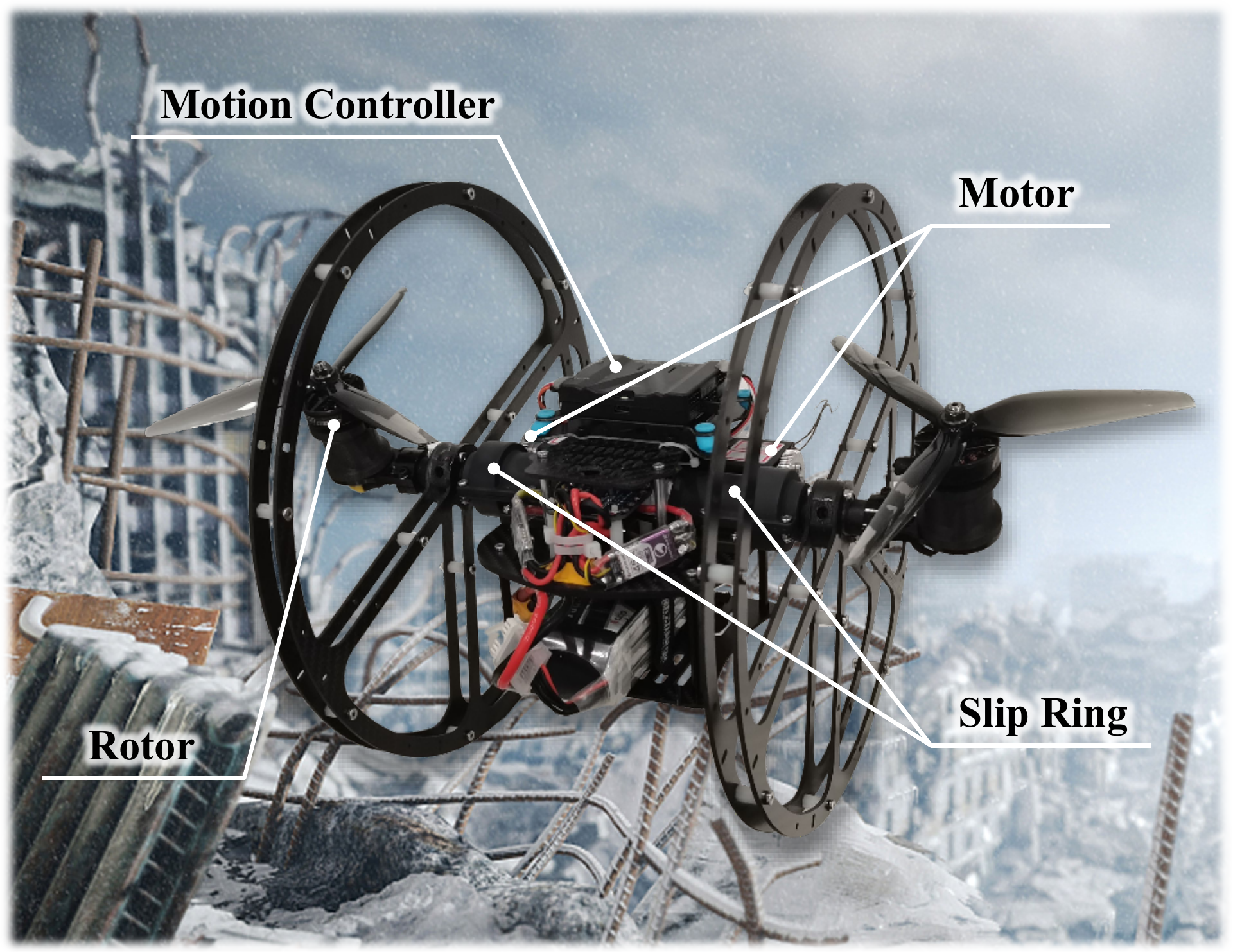}
	\end{center}
	\vspace{-0.2cm}
	\caption{
		\label{fig:abstract}
		Our CapsuleBot prototype.
	}
\end{figure}

\section{Introduction}
\IEEEPARstart{i}{N} RECENT years, Hybrid Aerial-Ground Robots have seen significant development in both academia , industry, and military applications \cite{sihite2023multi,zhang2022intelligent,israeldefense2023}. Hybrid aerial-ground robots integrate both flying and driving capabilities, offering significant advantages over ground robots in terms of cross-domain mobility and over aerial robots in terms of endurance. However, integrating these locomotion modes into a single robot prototype presents considerable design challenges.

Researchers have introduced numerous designs of aerial-ground robots to address this challenge, encompassing a wide array of combinations involving both flying and driving mechanisms. Firstly, in order to achieve ground mobility, combining the mechanisms of ground robots has become a common choice due to their well-developed maturity. To the best of the authors' knowledge, aerial-ground robots equipped with active wheels \cite{cao2023doublebee, kalantari2020drivocopter, tan2021multimodal, zheng2023roller, shi2024mtabot}, active walking mechanisms \cite{kim2021bipedal, sugihara2023design} and adaptive structure \cite{sihite2023multi, meiri2019flying, david2021design, morton2017small} typically necessitate additional motors compared to their flight configuration. This not only complicates the control system but also reduces the robot's reliability while increasing manufacturing and maintenance costs. In addition, aerial-ground robots equipped with passive mechanisms, such as passive cage \cite{kalantari2013design, kalantari2014modeling} or wheels \cite{zhang2023model, yang2022sytab, lin2024skater, qin2020hybrid, pan2023skywalker} do not require additional motors. However, the propulsion of these robots in ground mode is entirely derived from the thrust generated by the rotors, resulting in significant noise and dust, and impacting the control system due to the Wing-In-Ground effect \cite{matus2021ground}. 
The precise rolling angular conrol was also  difficult to achieve and not demonstrated by the scheme\cite{jia2023quadrolltor}.
In contrast, reusing specific motors from the flight configuration in ground mode shows promise in reducing the need for additional motors in aerial-ground robots. However, using rotors on a quadcopter as ground-actuated wheels would still require extra motors to facilitate the transition between flight and ground modes \cite{meiri2019flying, david2021design}. By analyzing common multicopter configurations, we find that the rotational direction of the motors used for tilting in a tilt-rotor multicopter could align with the rotation scheme of ground-actuated wheels. This suggests that reusing the tilting motors for wheel actuation could effectively address the ground mobility challenges faced by aerial-ground robots. Based on this, we design an an innovative mechanical structure named the actuated-wheel-rotor, which uses a servo motor and a brushless motor to function as both a tilting rotor in the air and an actuated wheel on the ground, it will be detailed in the Section \ref{sec: design}.

Then, regarding aerial mobility. Initially, in order to attain aerial mobility, multicopters are frequently selected for their vertical takeoff and landing capabilities. Among these, quadrotors are extensively employed because of their compact mechanism and straightforward dynamics \cite{sihite2023multi, israeldefense2023, kalantari2020drivocopter, tan2021multimodal, zheng2023roller, meiri2019flying, david2021design, morton2017small, kalantari2013design, kalantari2014modeling, zhang2023model, pan2023skywalker}. However, bi-copters generally exhibit greater energy efficiency \cite{qin2020gemini} and improved maneuverability in narrow spaces \cite{lin2024skater} compared to quadcopters of equivalent weight and size. And aerial-ground robots incorporating bi-copter have witnessed development in recent years \cite{cao2023doublebee, shi2024mtabot, yang2022sytab, lin2024skater, qin2020hybrid}.
However, existing aerial-ground bi-copter robots have not effectively addressed the challenges posed by additional motors and rotor-driven ground movement, which result in noise, dust, and complications in the control system. Fortunately, a bi-copter configuration equipped with two tilting motors can be enhanced with our actuated-wheel-rotor design. Therefore, we chose the bi-copter configuration for our flight design.

\begin{figure}[t]
	\begin{center}
		\includegraphics[width=1.0\columnwidth]{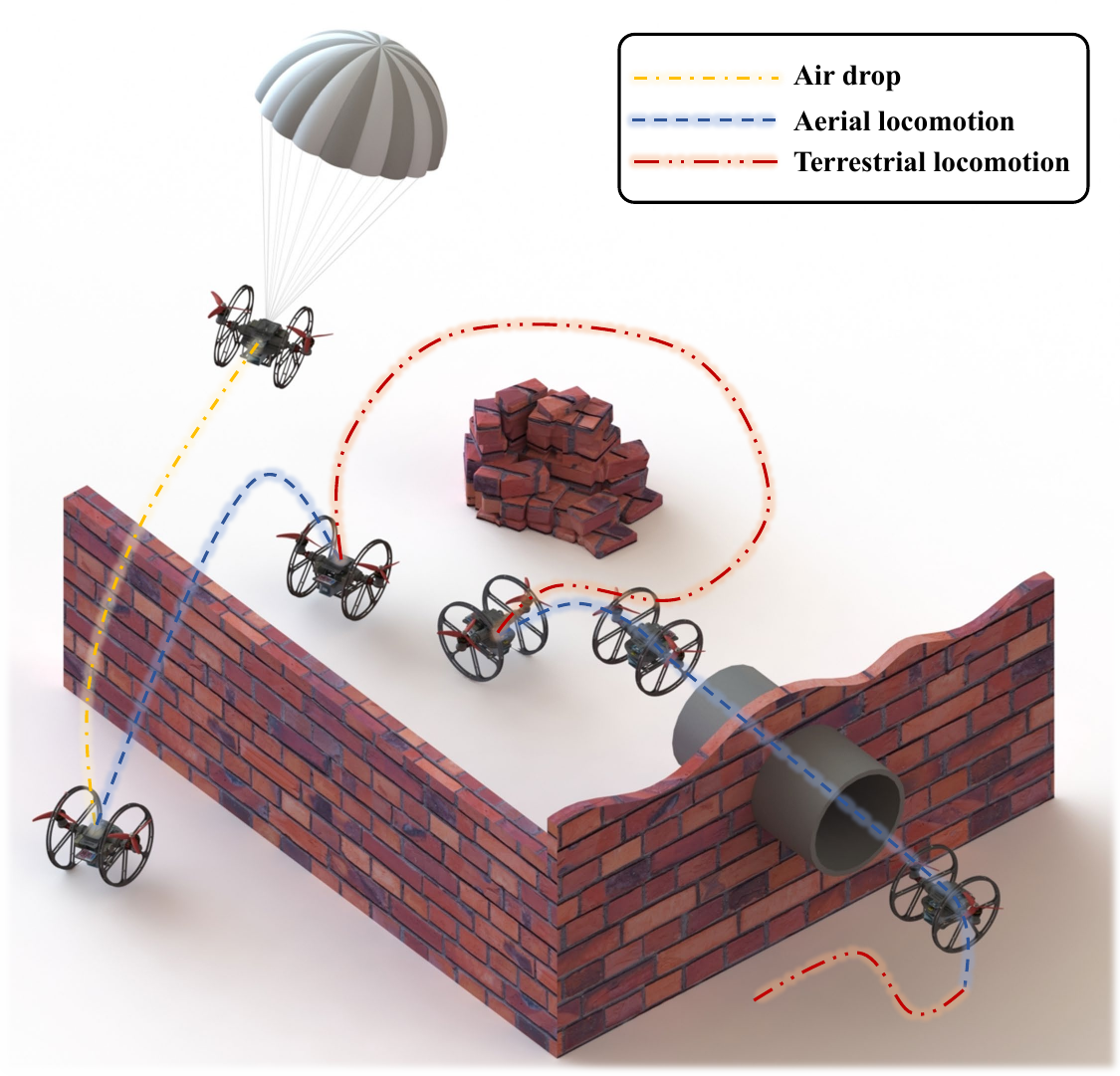}
	\end{center}
	\vspace{-0.2cm}
	\caption{
		\label{fig:application_scenario}
		An envisioned application scenario for CapsuleBot involves deploying the robot into a concealed reconnaissance area using either aerial or wheel modes. The aerial mode is utilized not only for air dropping and flying over obstacles that are difficult for ground mode to traverse but also for maneuvering through pipes and narrow spaces. Besides, ground mode is prioritized for tasks requiring low dust generation, low noise, and long endurance.
	}
\end{figure}

Based on the above analysis, we propose the design of CapsuleBot, a hybrid aerial-ground bi-copter robot. It combines the aerial maneuverability of a bi-copter with the low power consumption and low noise of a two-wheel self-balancing robot on the ground. This innovative design is centered around a novel mechanical structure called the actuated-wheel-rotor, which employs both a servo motor and a brushless motor to serve as a tilting rotor in the air and an actuated wheel on the ground. With two of these structures, CapsuleBot achieves hybrid motion using only four motors, eliminating the need for additional motors compared to a bi-copter. Additionally, we tackle the challenge of wire management during wheel mode by incorporating slip rings. Due to its long-endurance , low-noise, low-dust, and cross-domain mobility advantages, CapsuleBot is highly suitable not only for wildlife monitoring scenarios and archaeological sites but also for implementation in military or field environments, serving as a covert and persistent Intelligence, Surveillance, and Reconnaissance (ISR) platform.

In order to fully exploit the performance of CapsuleBot, we have developed comprehensive dynamics and control systems for both air and wheel modes, based on the bi-copter model and the two-wheel self-balancing robot model. A prototype of CapsuleBot has been constructed, and its performance in terms of low power consumption and low noise has been validated through experiments. Challenging tasks have demonstrated CapsuleBot's ability to climb steep inclines, fly over cliffs, and traverse rough terrains.

The main contributions of this paper are summarized as:
\begin{itemize}
	\item A novel hybrid aerial-ground bi-copter robot equipped with two actuated-wheel-rotors. The actuated-wheel-rotor utilizes a servo motor and a brushless motor to function as both a tilting rotor in the air and an actuated wheel on the ground.
	\item Comprehensive dynamics modeling and controller design encompassing both aerial and wheel modes.
	\item A series of real-world experiments and benchmark comparisons confirm the performance of CapsuleBot, demonstrating its high energy efficiency and low noise levels in wheel mode.
	\item Two challenging tasks demonstrate CapsuleBot’s capabilities in unstructured environments, including climbing steep inclines, flying over cliffs, and traversing rough terrains.
\end{itemize}

\section{Mechatronic system design}\label{sec: design}

\subsection{Design of Actuated-Wheel-Rotor}
Due to the alignment of the rotational direction of the motors used for tilting in a tilt-rotor multicopter and the rotation scheme of ground-actuated wheels, reusing the tilting motors for wheel actuation could solve the challenge of needing additional motors for ground mobility. Based on this concept, we propose an innovative mechanism called the actuated-wheel-rotor.

\begin{figure}[t]
	\begin{center}
		\includegraphics[width=1.0\columnwidth]{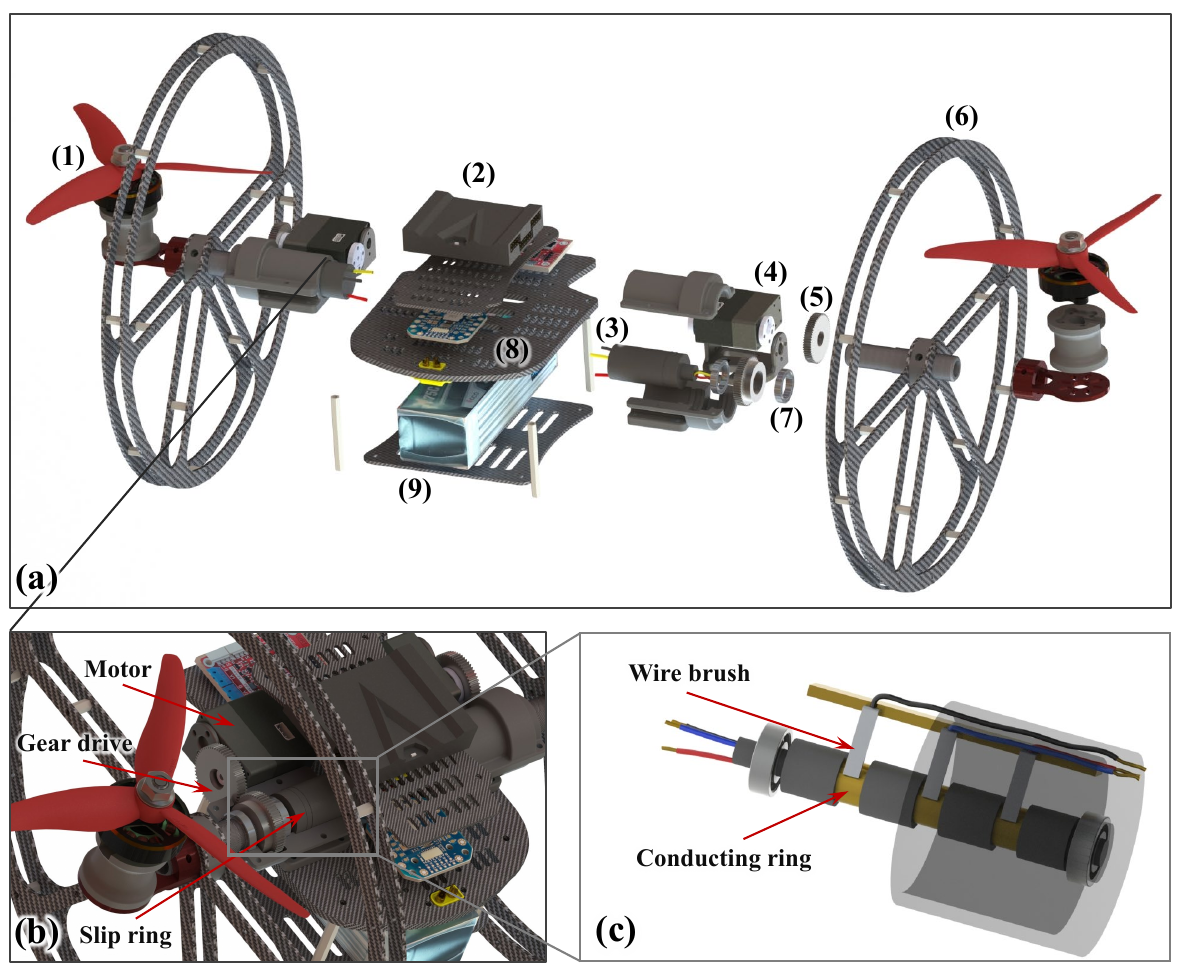}
	\end{center}
	\vspace{-0.2cm}
	\caption{
		\label{fig:system_architecture_and_components}
		The composition of CapsuleBot is detailed below. (a) The comprehensive structural diagram of the robot. (b) Details of gear transmission structures. (c) Details of a slip ring. The serial numbers on (a) correspond to specific components: (1) actuated-wheel-rotor, (2) flight controller and electronic speed controller (ESC), (3) slip ring, (4) servo motor, (5) gear, (6) carbon fiber wheel, (7) bearing, (8) carbon fiber frame, (9) battery.
	}
\end{figure}

The structure of actuated-wheel-rotor is illustrated in Fig. \ref{fig:system_architecture_and_components}. 
Each actuated-wheel-rotor comprises a brushless motor, a servo motor, a slip ring, a wheel and transmission mechanisms. The wheel consists of two hollow \( 1.5 \, \mathrm{mm} \) carbon plates connected by nylon columns, striking a balance between weight and rigidity. The brushless motor (T-MOTOR F90 \( 1500 \, \mathrm{KV} \)) and 7-inch three-blade propeller (HQ70403) provide thrust in flight mode.
It is important to note that the rotor only operates in aerial mode and remains inactive in wheel mode. The servo motor (Feetech STS3025BL) dual-functions as both the wheel rotation and rotor tilt driver, aiming to minimize the number of motors. Incorporating slip rings addresses the issue of brushless motor wire entanglement caused by continuous rotation in wheel mode as shown in Fig. \ref{fig:system_architecture_and_components}(c). 
We selected the slip ring is rated at \( 30 \, \mathrm{A} \) per channel, with three channels providing a total of \( 90 \, \mathrm{A} \), which is sufficient for the normal flying of our robot.
In this design, the slip ring are required to be coaxial with the drive shaft of the carbon fiber wheel, which also serves as the rotor tilt axis. Therefore, directly driving this shaft with the servo motor would result in spatial positioning conflicts, necessitating the design of corresponding transmission structures as shown in Fig. \ref{fig:system_architecture_and_components}(b). The torque from the servo motor is then transmitted through the gear transmission structures, with a transmission ratio of $i = 1:1$.

\subsection{System Architecture and Components}
\begin{figure}[t]
	\begin{center}
		\includegraphics[width=1.0\columnwidth]{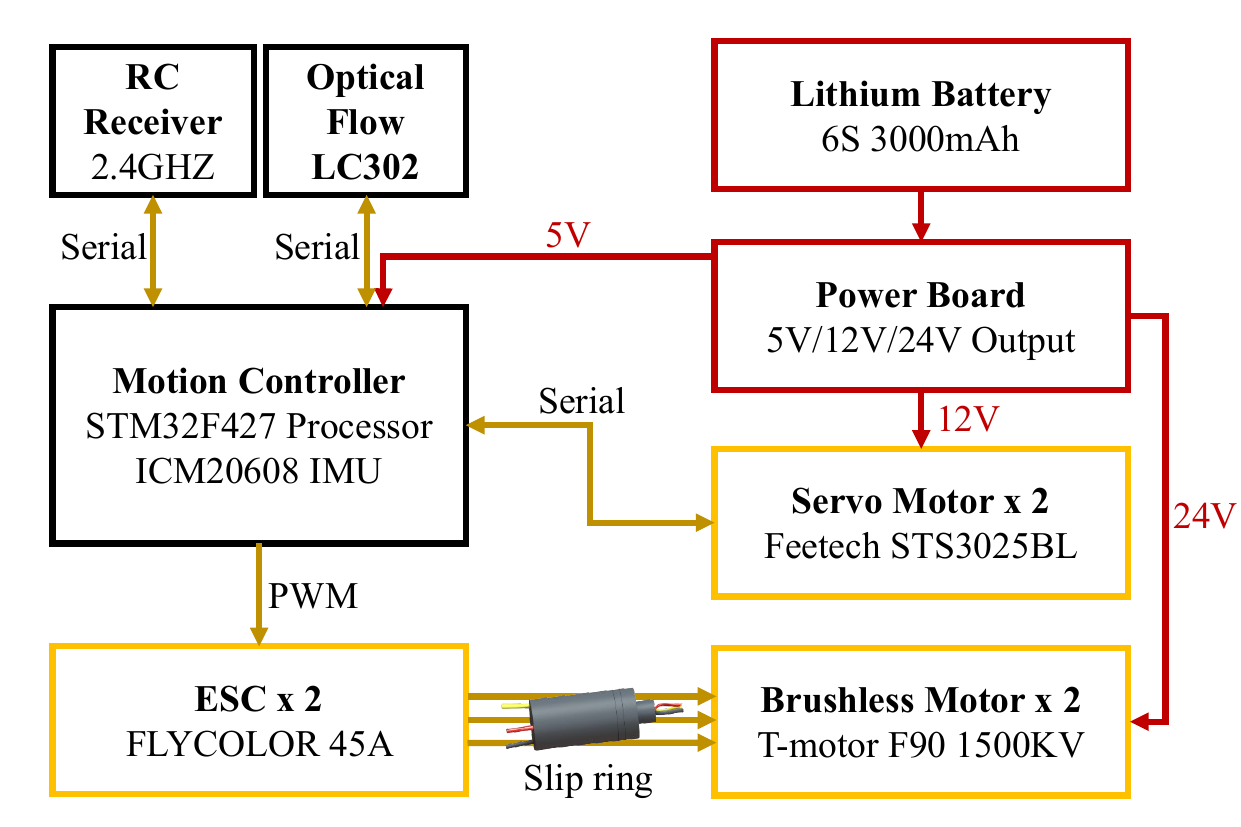}
	\end{center}
	\caption{
		\label{fig:electrical_architecture}
		Electrical architecture of CapsuleBot, where the yellow lines represent logic circuits, and the red lines denote power circuits.
	}
\end{figure}

CapsuleBot consists of a main body and two actuated-wheel-rotors arranged in a symmetrical layout.The structure is illustrated in Fig. \ref{fig:system_architecture_and_components}(a). The robot's main body is also designed with axial symmetry and constructed using a carbon fiber frame. The total mass of CapsuleBot is approximately \( 1.8 \, \mathrm{kg} \). Table. \ref{table:specifications} lists the model and weight of key components. The distance between the two rotors is about \( 30 \, \mathrm{cm} \), and width of robot is close to \( 48 \, \mathrm{cm} \). The diameter of carbon fiber wheels and the hight/length of robot are about \( 28 \, \mathrm{cm} \). Fig. \ref{fig:electrical_architecture} shows CapsuleBot’s electrical architecture. The system is powered by a \( 3000 \, \mathrm{mAh} \) \( 6 \, \mathrm{s} \) \( 75 \, \mathrm{c} \) lithium battery. For multimodal motion control, we employ a microcontroller featuring a STM32F427 ARM processor and an integrated IMU chip (ICM20608). CapsuleBot is controlled by a remote controller, and Kalman filtering is conducted to update its orientation at \( 400 \, \mathrm{Hz} \).

\begin{table}[!htb]  
	\renewcommand\arraystretch{1.3}
	\centering
	\caption{Model and weight of key components}  
	\label{table:specifications} 
	\begin{tabular*}{\linewidth}{@{}l|l|l@{}} 
		\toprule[2pt]
		\textbf{Component} & \textbf{Model} & \textbf{Weight(g)} \\ \midrule  
		Battery & \( 3000 \, \mathrm{mAh} \) \( 6 \, \mathrm{s} \) \( 75 \, \mathrm{c} \) & 480  \\ 
		Servo motors & Feetech STS3025BL & 89  \\      
		Brushless motors & T-Motor F90 \( 1500 \, \mathrm{KV} \) & 46.6  \\ 
		Motion controller & Facinnov Mcontroller V7 & 25.6  \\ 
		ESC & FLYCOLOR \( 45 \, \mathrm{A} \) & 10.2  \\ 
		Propellers & HQPROP HQ70403 & 9  \\ 
		Wheels & Carbon plates and nylon columns & 37.3  \\ \bottomrule[2pt]   
	\end{tabular*}   
\end{table}

\section{MODELING AND CONTROL}\label{sec: modelling}
In CapsuleBot, the dynamics models consist of two parts, the aerial and wheel modes (the coordinate frames are shown in Fig. \ref{fig:dynamic_model} for later use). The wheel mode further include terrestrial and transitional states.

\subsection{Aerial Mode}
\subsubsection{MODELING}
\begin{figure}[t]
	\begin{center}
		\includegraphics[width=1.0\columnwidth]{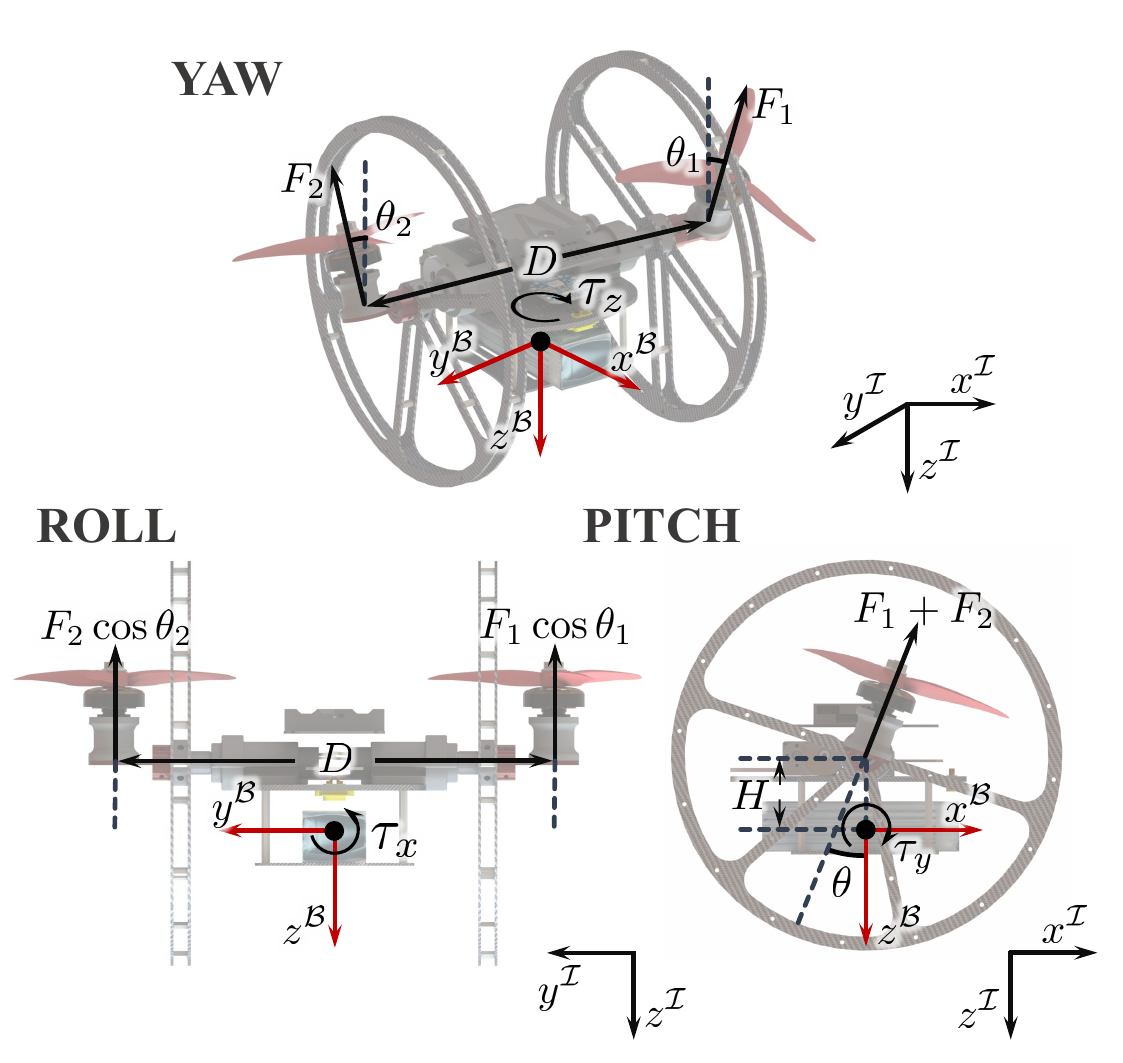}
	\end{center}
	\vspace{-0.2cm}
	\caption{
		\label{fig:dynamic_model}
		Dynamic model.
	}
\end{figure}

When CapsuleBot is in aerial mode, there are no frictional or contact forces acting on its wheels. Consequently, the dynamic equations can be simplified to those of a standard bi-copter \cite{qin2020gemini}. The torque generated is produced by the rotor thrusts:
\begin{equation}
	\label{eq:standard_bi-copter}
	\begin{cases}
		\tau_x=\left(F_1 \cos \theta_1 - F_2 \cos \theta_2\right) D/2,\\
		\tau_y=(F_1 \sin \theta_1+F_2 \sin  \theta_2) H, \\
		\tau_z=\left(F_1 \sin \theta_1 - F_2 \sin \theta_2\right) D/2 .
	\end{cases}
\end{equation}
Where $F_1$, $F_2$, $\theta_1$, and $\theta_2$ represent the thrust generated by the two rotors and the angle between them and the $\boldsymbol{z}^{\mathcal{B}}$ axis, as shown in Fig. \ref{fig:dynamic_model}. The inertial frame $\left(\boldsymbol{x}^{\mathcal{I}}, \boldsymbol{y}^{\mathcal{I}}, \boldsymbol{z}^{\mathcal{I}}\right)$ and body frame $\left(\boldsymbol{x}^{\mathcal{B}}, \boldsymbol{y}^{\mathcal{B}}, \boldsymbol{z}^{\mathcal{B}}\right)$ are defined. Throughout the paper, the superscripts ${ }^{\mathcal{I}}$ and ${ }^{\mathcal{B}}$ will be used to denote the inertial and body frame, respectively. The inertial frame follows the convention of North-East-Down (NED). The dynamics of the aerial mode adhere to a standard rigid motion model:
\begin{subequations}
	\begin{equation}
		\label{eq:standard_rigid_motion_model_1}
		\setlength{\arraycolsep}{1.2pt}
		{\left[\begin{array}{cc}
				m \boldsymbol{I} & 0 \\
				0 & \boldsymbol{J}^{\mathcal{B}}
			\end{array}\right]\left[\begin{array}{c}
				\dot{\boldsymbol{v}}^{\mathcal{I}} \\
				\dot{\boldsymbol{\Omega}}^{\mathcal{B}}
			\end{array}\right]+\left[\begin{array}{c}
				0 \\
				\widehat{\boldsymbol{\Omega}}^{\mathcal{B}} \boldsymbol{J}^{\mathcal{B}} \boldsymbol{\Omega}^{\mathcal{B}}
			\end{array}\right]=\left[\begin{array}{c}
				\boldsymbol{f}_g \\
				0
			\end{array}\right]+\left[\begin{array}{c}
				\boldsymbol{R} \boldsymbol{f}_F^{\mathcal{B}} \\
				\boldsymbol{\tau}^{\mathcal{B}}
			\end{array}\right]},
	\end{equation}
	\begin{equation}
		\label{eq:standard_rigid_motion_model_2}
		\boldsymbol{R}=\boldsymbol{R}_z(\kappa) \boldsymbol{R}_y(\omega) \boldsymbol{R}_x(\varphi).
	\end{equation}
\end{subequations}
In Eq. \ref{eq:standard_rigid_motion_model_1} and \ref{eq:standard_rigid_motion_model_2}, the variables $m$, $\boldsymbol{I}$, and $\boldsymbol{J}^{\mathcal{B}}$ represent the mass, the identity matrix in $\mathbb{R}^{3 \times 3}$, and the inertia matrix, respectively. The angular velocity vector $\boldsymbol{\Omega}^{\mathcal{B}}$ is represented in the body frame, while $\widehat{\boldsymbol{\Omega}}^{\mathcal{B}}$ is the skew-symmetric cross product matrix of $\boldsymbol{\Omega}^{\mathcal{B}}$. The velocity vector $\boldsymbol{v}^{\mathcal{I}}$ is represented in the inertial frame. The gravity vector $\boldsymbol{f}_g=\left[\setlength{\arraycolsep}{1.2pt}\begin{array}{lll}0 & 0 & mg\end{array}\right]^T$ represents gravity in the inertial frame. The rotation matrix $\boldsymbol{R}$ transforms from the inertial frame to the body frame, following the Z-Y-X Tait-Bryan order, where $\kappa$, $\omega$, and $\varphi$ represent the yaw, pitch, and roll Euler angles, respectively. The moment vector produced by the rotors in the body frame is denoted as $\boldsymbol{\tau}^{\mathcal{B}}=\left[\setlength{\arraycolsep}{1.2pt}\begin{array}{lll}\tau_x & \tau_y & \tau_z\end{array}\right]^T$, where $\tau_x$, $\tau_y$, and $\tau_z$ are detailed in Eq. \ref{eq:standard_bi-copter}. The force vector $\boldsymbol{f}_F^{\mathcal{B}}=\left[\setlength{\arraycolsep}{1.2pt}\begin{array}{lll}0 & 0 & F\end{array}\right]^T$, where $F=F_1 \cos \theta_1+F_2 \cos \theta_2$.

\subsubsection{CONTROL}
\begin{figure}[t]
	\begin{center}
		\includegraphics[width=1.0\columnwidth]{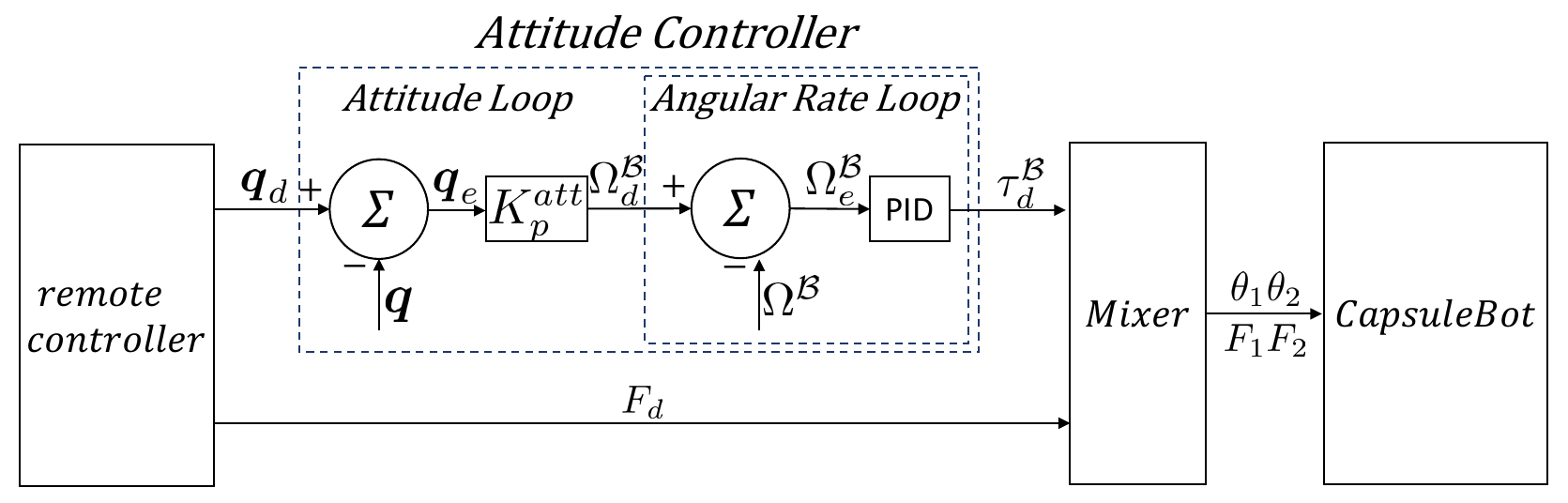}
	\end{center}
	\vspace{-0.2cm}
	\caption{
		\label{fig:control_structure}
		Control structure of aerial mode.
	}
\end{figure}

The control structure of the aerial mode is depicted in Fig. \ref{fig:control_structure}. The remote control generates the desired attitude, denoted as $\boldsymbol{q}_d$, and the force along the body's $\mathrm{Z}$ axis, referred to as $F_d$. The attitude controller calculates the desired moment, $\tau_d^{\mathcal{B}}$, using the desired attitude. Finally, the mixer determines the motor thrust commands and servo angle commands based on $\tau_d^{\mathcal{B}}$ and $F_d$.

Attitude Controller: The attitude controller is a cascaded controller consisting of an outer loop, referred to as the angular loop. This loop utilizes a proportional controller to track the desired attitude $\boldsymbol{q}_d$. The attitude is represented using a quaternion $(\boldsymbol{q}=[\eta, \boldsymbol{\epsilon}])$, where $\boldsymbol{\epsilon}$ and $\eta$ represent the vector and scalar parts of the quaternion, respectively. The quaternion representation enables the calculation of the attitude error $\boldsymbol{q}_e$ and the desired angular velocity $\boldsymbol{\Omega}_d^{\mathcal{B}}$ using the ``Quaternion linear" method described in \cite{lyu2017hierarchical}:
\begin{subequations}
	\begin{equation}
			\boldsymbol{q}_e=\boldsymbol{q}_d{ }^* \otimes \boldsymbol{q}=\left[\eta_e \boldsymbol{\epsilon}_e^T\right]^T,
	\end{equation}
	\begin{equation}
		\varphi=2 \cdot \operatorname{atan} 2\left(\left\|\epsilon_{\mathrm{e}}\right\|, \eta_{\mathrm{e}}\right),
	\end{equation}
	\begin{equation}
		\boldsymbol{\Omega}_d^{\mathcal{B}}=K_p^{att} \cdot \boldsymbol{q}_e^{\mathcal{B}}=K_p^{att} \cdot \operatorname{sign}\left(\eta_{\mathrm{e}}\right) \frac{\varphi}{\sin \left(\frac{\varphi}{2}\right)} \boldsymbol{\epsilon}.
	\end{equation}
\end{subequations}
The desired attitude and actual attitude are represented by $\boldsymbol{q}_d$ and $\boldsymbol{q}$, respectively. The gain from attitude error $\boldsymbol{q}_e^{\mathcal{B}}$ to desired angular velocity $\boldsymbol{\Omega}_d^{\mathcal{B}}$ is denoted as $K_p^{att}$. The inner loop, also known as the angular rate loop, utilizes a PID controller (Eq. \ref{eq:PID_controller}) to track the desired angular velocity $\boldsymbol{\Omega}_d^{\mathcal{B}}$:
\begin{subequations}
	\label{eq:PID_controller}
	\begin{equation}
			\boldsymbol{\Omega}_e^{\mathcal{B}}=\boldsymbol{\Omega}_d^{\mathcal{B}}-\boldsymbol{\Omega}^{\mathcal{B}},\\
	\end{equation}
	\begin{equation}
		\boldsymbol{\tau}_d^{\mathcal{B}}=K_p^{rt} \cdot \boldsymbol{\Omega}_e^{\mathcal{B}}+K_i^{rt} \cdot \int \boldsymbol{\Omega}_e^{\mathcal{B}}+K_d^{rt} \cdot \dot{\boldsymbol{\Omega}}_e^{\mathcal{B}}.
	\end{equation}
\end{subequations}
The error in angular velocity and the current angular velocity are denoted as $\boldsymbol{\Omega}_e^{\mathcal{B}}$ and $\boldsymbol{\Omega}^{\mathcal{B}}$, respectively. The gains of the PID terms are represented by $K_p^{rt}, K_i^{rt}$, and $K_d^{rt}$. The output of the inner loop is the desired moment $\boldsymbol{\tau}_d^{\mathcal{B}}=\left[\setlength{\arraycolsep}{1.2pt}\begin{array}{lll}\tau_{x d} & \tau_{y d} & \tau_{z d}\end{array}\right]^T$.

Mixer: The relationship between moments ($\tau_x, \tau_y, \tau_z$), body-Z force ($F$), and actuator output ($\theta_1, \theta_2, F_1, F_2$) is illustrated in Fig. \ref{fig:control_structure}:
\begin{equation}
	\label{eq:mixer}
	\setlength{\arraycolsep}{1.5pt}
	\left[\begin{array}{cccc}
		0 & 1 & 0 & 1 \\
		0 & D/2 & 0 & -D/2 \\
		H & 0 & H & 0 \\
		D/2 & 0 & -D/2 & 0
	\end{array}\right]\left[\begin{array}{c}
		F_1 \sin \theta_1 \\
		F_1 \cos \theta_1 \\
		F_2 \sin \theta_2 \\
		F_2 \cos \theta_2
	\end{array}\right]=\left[\begin{array}{c}
		F \\
		\tau_x \\
		\tau_y \\
		\tau_z
	\end{array}\right].
\end{equation}
Given the desired moments $\boldsymbol{\tau}_d^{\mathcal{B}}=\left[\tau_{x d}, \tau_{y d}, \tau_{z d}\right]^T$ and the desired body-Z force $F_d$, the actuator outputs $\theta_1$, $\theta_2$, and the rotor thrust outputs $F_1$, $F_2$ can be solved from Eq. \ref{eq:mixer} as follows:
\begin{subequations}
	\begin{equation}
		\theta_1=\operatorname{atan}\left(\frac{D \cdot \tau_{y d} + 2 \cdot H \cdot \tau_{z d}}{H \cdot D \cdot F_d + 2 \cdot H \cdot \tau_{x d}}\right),
	\end{equation}
	\begin{equation}
		\theta_2=\operatorname{atan}\left(\frac{D \cdot \tau_{y d}-2 \cdot H \cdot \tau_{z d}}{H \cdot D  \cdot F_d - 2 \cdot H \cdot \tau_{x d}}\right),
	\end{equation}
	\begin{equation}
		F_1=\frac{1}{2} \cdot \sqrt{\left(\frac{\tau_{y d}}{H}+\frac{\tau_{z d}}{D/2}\right)^2+\left(F_d+\frac{\tau_{z d}}{D/2}\right)^2},
	\end{equation}
	\begin{equation}
		F_2=\frac{1}{2} \cdot \sqrt{\left(\frac{\tau_{x d}}{H}-\frac{\tau_{z d}}{D/2}\right)^2+\left(F_d-\frac{\tau_{y d}}{D/2}\right)^2}.
	\end{equation}
\end{subequations}

\subsection{Wheel Mode}
CapsuleBot's control scheme in Wheel mode is similar to that of a balance bot.
\subsubsection{TERRESTRIAL STATE}
It employs a cascade PID controller, with the inner loop tracking the desired pitch angle of the robot \cite{cao2023doublebee, Philip2020balance} and the outer loop tracking the desired velocity along the body's $x$-axis.
\begin{subequations}
	\begin{equation}
		\delta_e=K_v^{terr}\left(v_{desired}-v\right)-\delta,
	\end{equation}
	\begin{equation}
		\omega_{whl 1}=K_p^{terr} \delta_e+K_i^{terr} \int \delta_e+K_d^{terr} \dot{\delta}_e-K_\gamma^{terr}\left(\omega_{\gamma d}-\dot{\gamma}\right),
	\end{equation}
	\begin{equation}
		\omega_{whl 2}=K_p^{terr} \delta_e+K_i^{terr} \int \delta_e+K_d^{terr} \dot{\delta}_e+K_\gamma^{terr}\left(\omega_{\gamma d}-\dot{\gamma}\right).
	\end{equation}
\end{subequations}
The variables $\delta$, $v_{desired}$, $v$, $\omega_{\gamma d}$, and $\dot{\gamma}$ represent actual values of the pitch angle under terrestrial state, the expected and actual values of the vehicle velocity, and the steer (yaw) rate, respectively. The gain of the velocity control loop is denoted as $K_v^{terr}$, and the inner loop PID terms are defined as $K_p^{terr}$, $K_i^{terr}$, $K_d^{terr}$, and $K_\gamma^{terr}$. Additionally, $\omega_{whl 1}$ and $\omega_{whl 2}$ represent the angular velocity applied by the motors on the wheels.
\subsubsection{TRANSITIONAL STATE}
The transitional state refers to the phase during which CapsuleBot transitions from rolling to flying, with the objective of orienting the two rotors vertically upward. It employs a cascade PID controller, where the inner loop tracks the desired pitch angle of zero to maintain balance, and the outer loop tracks the desired wheel position to ensure the wheels remain at the encoder's zero point, thereby achieving the vertical orientation of the two rotors.

Balance Controller:
\begin{subequations}
	\begin{equation}
		\delta_{desired} = 0,
	\end{equation}
	\begin{equation}
		\delta_e = \delta_{desired} - \delta,
	\end{equation}
	\begin{equation}
		u_{\delta} = K_p^{bal} \delta_e + K_i^{bal} \int \delta_e \, dt + K_d^{bal} \frac{d \delta_e}{dt}.
	\end{equation}
\end{subequations}
The variables $\delta_{desired}$ represent the expected pitch angle values during the transitional state, set to zero to maintain balance. The gain the inner loop PID terms are defined as $K_p^{bal}$, $K_i^{bal}$ and $K_d^{bal}$.

Wheel Position Controller:
\begin{subequations}
	\begin{equation}
		x_{desired} = 0,
	\end{equation}
	\begin{equation}
		x_e = x_{desired} - x,
	\end{equation}
	\begin{equation}
		u_x = K_p^{pos} e_x + K_i^{pos} \int e_x \, dt + K_d^{pos} \frac{d e_x}{dt}.
	\end{equation}
\end{subequations}
The variables $x_{desired}$ represent the expected wheel position values during the transitional state, set to zero to achieving the vertical orientation of rotor. The gain the outer loop PID terms are defined as $K_p^{pos}$, $K_i^{pos}$ and $K_d^{pos}$.

Signal Combiner:
\begin{equation}
	\omega_{whl} = k_{\delta} \cdot u_{\delta} + k_{x} \cdot u_{x}.
\end{equation}
The variables $k_{\delta}$ and $k_{x}$ are weight coefficients used to adjust the contributions of the balance controller and wheel position controller to the final motor control signal. $\omega_{whl}$ represents the angular velocity applied by the motors to the wheels.

\section{EXPERIMENTAL VALIDATION}\label{sec: exp}
Fig. \ref{fig:abstract} showcases the platform used for conducting real-world experiments.

\subsection{Energy Efficiency Validation}
In this experiment, our main objective is to evaluate the energy consumption of CapsuleBot. In wheel mode, the robot is programmed to roll at a velocity of approximately \( 0.5 \, \mathrm{m/s} \) with a circular trajectory radius of about \( 1 \, \mathrm{m} \). The energy consumption of the robot in aerial mode is tested during manual control, with a flight speed of approximately \( 0.5 \, \mathrm{m/s} \), which is roughly the same as in wheel mode.

We referred to Shi's work \cite{shi2024mtabot}, where the power efficiency, considering the robot's mass, was evaluated using the weight-to-power ratio ($g/W$), as shown in Table. \ref{table:1}. The mean power of aerial mode, denoted as $P_a$, is about \( 480.6 \, \mathrm{W} \) (with a weight-to-power ratio of \( 3.7 \, \mathrm{g/W} \), denoted as $M/P_a$), while the mean power of wheel mode, denoted as $P_g$, is about \( 5.2 \, \mathrm{W} \) (with a weight-to-power ratio of \( 346.2 \, \mathrm{g/W} \), denoted as $M/P_g$). Therefore, by combining Pan's \cite{pan2023skywalker} and Shi's work \cite{shi2024mtabot}, we calculate the comprehensive efficiency as:
\begin{equation}
	\eta=\left(1-\frac{P_g}{P_a}\right) \times 100 \% = \left(1-\frac{M/P_a}{M/P_g}\right) \times 100 \% =98.9 \%.
\end{equation}

We conduct a comparative analysis of CapsuleBot's average power consumption with three typical aerial-ground bi-copter robots: Cao's, which employs double active wheels \cite{cao2023doublebee}; Yang's, which utilizes double passive wheels \cite{yang2022sytab}; and Qin's, which is a single passive wheel-based robot \cite{qin2020hybrid}. Table. \ref{table:1} presents the power consumption across various operational modes. This comparison provides insights into the energy efficiency and operational characteristics of CapsuleBot relative to existing bi-copter designs, highlighting its potential advantages in practical applications.

\begin{table}[!htb] 
	\scriptsize
	\renewcommand\arraystretch{1.3}
	\centering
	\caption{Power/Efficiency of air-ground bi-copter robots.}  
	\label{table:1} 
	\begin{tabular*}{\linewidth}{@{}l|l|l|l|l@{}} 
		\toprule[2pt]
		\textbf{Power / Efficiency}&\textbf{Proposed} & \textbf{Cao's\cite{cao2023doublebee}} & \textbf{Yang's\cite{yang2022sytab}}& \textbf{Qin's\cite{qin2020hybrid}}\\ \midrule  
		Ground power ($W$) & 5.2 & 10 & 19 & 106  \\ 
		Ground efficiency ($g/W$) & 346.2 & 278.0 & 74.3 & 18.4  \\ 
		Aerial power ($W$)& 480.6 & 462 & 290 & 454.5  \\  
		Aerial efficiency ($g/W$) & 3.7 & 6.0 & 4.9 & 4.3  \\
		Comprehensive efficiency & 98.9$\%$ & 97.8$\%$ & 93.4$\%$ & 76.7$\%$  \\ \bottomrule[2pt]
	\end{tabular*} 
\end{table}

\subsection{Noise Level}
Two series of experiments being conducted to evaluate the noise performance in wheel mode and aerial mode. 

\begin{figure}[t]
	\begin{center}
		\includegraphics[width=1.0\columnwidth]{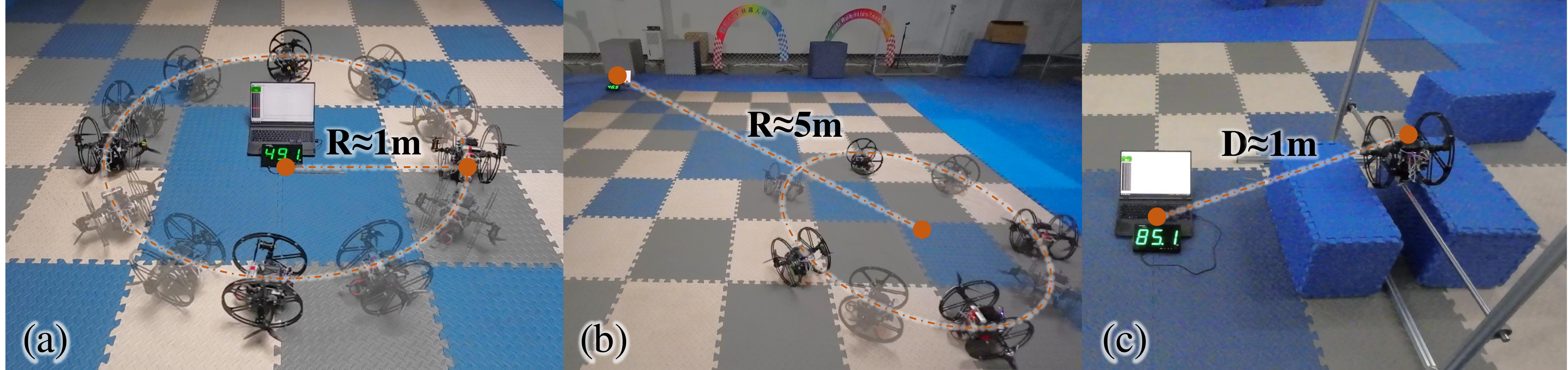}
	\end{center}
	\vspace{-0.2cm}
	\caption{
		\label{fig:noise level experiment2}
		(a) Wheel mode (\( 1 \, \mathrm{m} \)). (b) Wheel mode (\( 5 \, \mathrm{m} \)). (c) Aerial mode (\( 1 \, \mathrm{m} \)). Experiments under ambient noise conditions of approximately \( 42.9 \, \mathrm{dB} \).
	}
\end{figure}

\begin{figure}[t]
	\begin{center}
		\includegraphics[width=1.0\columnwidth]{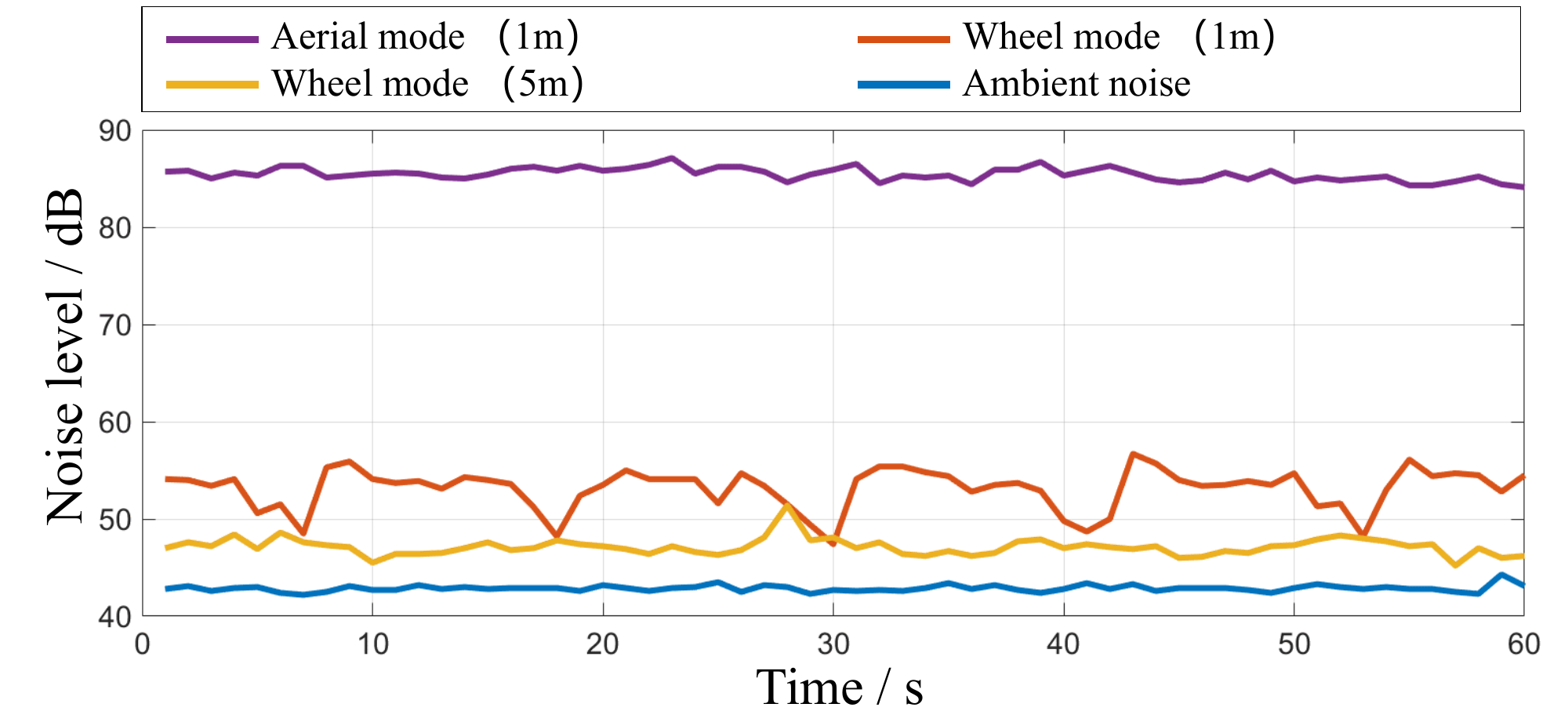}
	\end{center}
	\vspace{-0.2cm}
	\caption{
		\label{fig:noise level experiment}
		Results of noise level experiments.
	}
\end{figure}

In the noise level experiments of wheel mode, as shown in Fig. \ref{fig:noise level experiment2} (a) and (b), the robot is programmed to roll at a velocity of approximately \( 0.5 \, \mathrm{m/s} \) with a circular trajectory radius of about \( 1 \, \mathrm{m} \). The sound level monitor is positioned approximately \( 1 \, \mathrm{m} \) or \( 5 \, \mathrm{m} \) away from the robot. Data collection is continuous throughout the experiment, and the specific experimental data can be seen in Fig. \ref{fig:noise level experiment}, with recorded values ranging from a minimum of \( 47.4 \, \mathrm{dB} \) to a maximum of \( 56.7 \, \mathrm{dB} \), and an average of \( 53.1 \, \mathrm{dB} \). The noise generated from the motion in wheel mode, which is approximately \( 5 \, \mathrm{m} \) away, averages \( 47.1 \, \mathrm{dB} \).

In the noise level experiments of aerial mode, as shown in Fig. \ref{fig:noise level experiment2} (c), the robot is connected to the flight control test rack to maintain a stable distance of \( 1 \, \mathrm{m} \) between the sound level monitor and the robot more effectively. It begins in an idle state and is manually maneuvered to maintain the throttle at a fluctuation of 45\% to 50\% to simulate the actual aerial mode. Throughout the entire duration of the experiment, the instrument consistently records data, capturing values that range from a minimum of \( 84.1 \, \mathrm{dB} \) to a maximum of \( 87.1 \, \mathrm{dB} \), with an average of \( 85.4 \, \mathrm{dB} \). 

The experimental results depicted in Fig. \ref{fig:noise level experiment} reveal significant insights into CapsuleBot's noise levels during different operational modes. In wheel mode, the robot's noise level increases by only 23.9\% compared to ambient noise levels, and it decreases by 44.8\% compared to its noise output in aerial mode at a distance of \( 1 \, \mathrm{m} \). At a distance of \( 5 \, \mathrm{m} \), the sound produced by CapsuleBot during ground operation is barely perceptible and does not significantly differ from ambient noise levels. 
To the best knowledge of authors, the noise level of hybrid terrestrial/aerial vehicles has not yet been evaluated in the field. Compared to passive wheel-based robots, which rely entirely on rotors and generate significant noise, active wheel-based robots generally have advantages in terms of noise level in most ground situations. These findings highlight CapsuleBot's efficient noise management capabilities, particularly in ground-based operations, making it suitable for scenarios requiring low acoustic signature and stealthy operation.

\begin{figure}[t]
	\begin{center}
		\includegraphics[width=1\columnwidth]{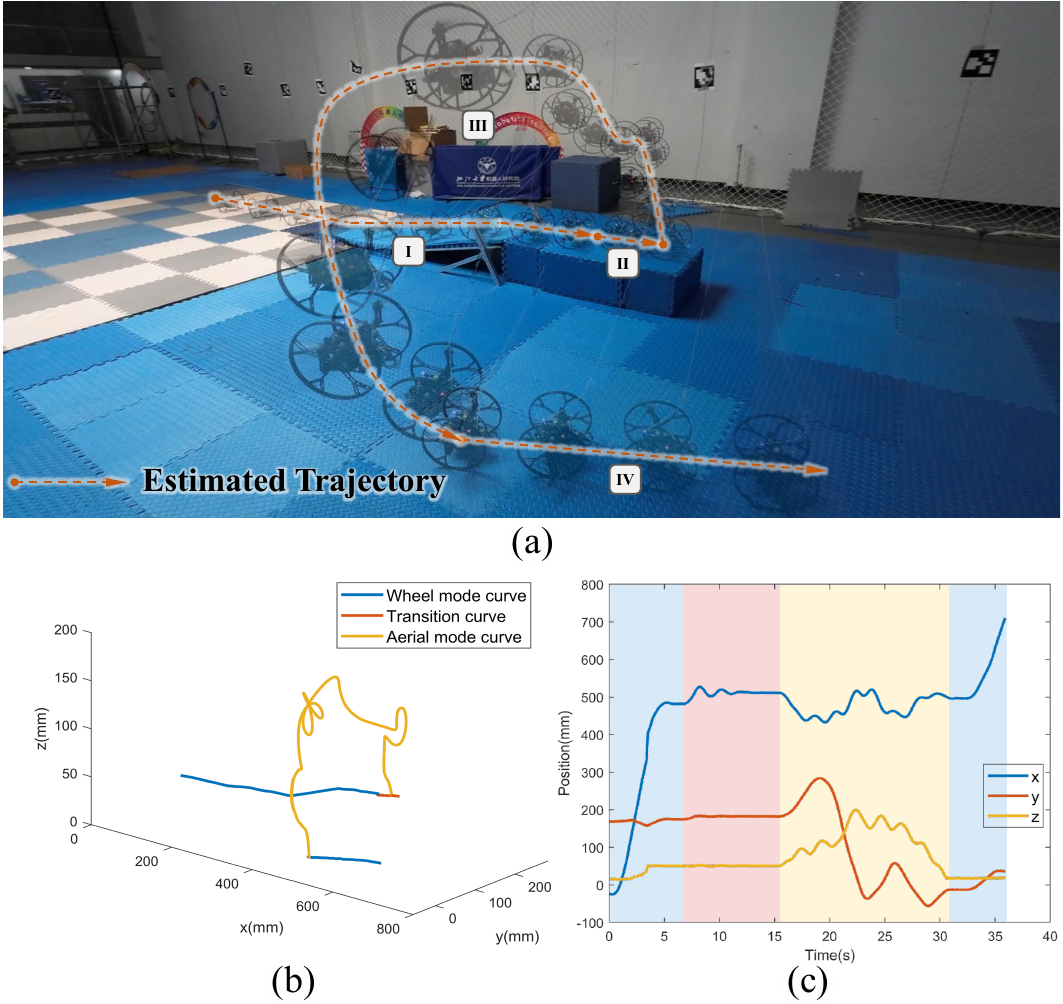}
	\end{center}
	\vspace{-0.2cm}
	\caption{
		\label{fig:first_task}
		(a) Real-world experiments, including climbing steep inclines, transitioning, flying over cliffs and keep rolling. (b) 3D trajectory points corresponding to (a). (c) Position response corresponding to (a).
	}
\end{figure}

\begin{figure}[t]
	\begin{center}
		\includegraphics[width=1\columnwidth]{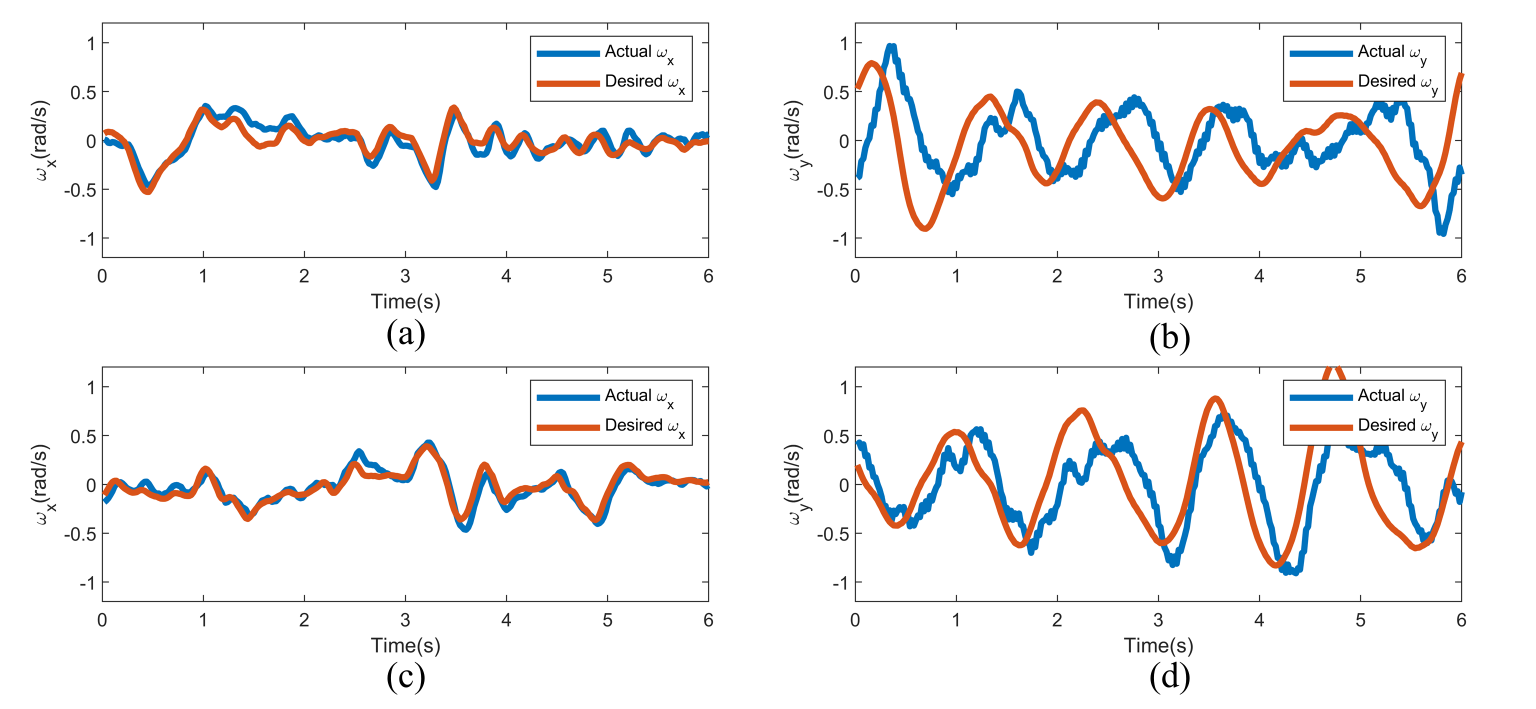}
	\end{center}
	\caption{
		\label{fig:manual_flight}
		Angular velocity tracking of CapsuleBot with wheel (a,b) or without wheel (c,d) during the manual flight.
	}
\end{figure}

\subsection{Challenging Tasks}
We conduct two challenging tasks based on the structural characteristics and application prospects of CapsuleBot: climbing steep inclines, transitioning, flying over cliffs, and traversing rough terrain.

\subsubsection{CLIMB STEEP AND FLY OVER CLIFFS}
In the first task scenario (Fig. \ref{fig:first_task}): Trajectory \uppercase\expandafter{\romannumeral1} shows the robot successfully ascending a steep incline, the maximum slope angle is about \( 10^\circ \). Trajectory \uppercase\expandafter{\romannumeral2} means that upon reaching the elevated platform, the robot smoothly adjusts its orientation, positioning the two rotors vertically upward in preparation for aerial mode using the control scheme detailed in Section \uppercase\expandafter{\romannumeral3}-transitional state. Trajectory \uppercase\expandafter{\romannumeral3} illustrates the robot's flight over cliffs in aerial mode, demonstrating its ability to traverse challenging airborne maneuvers. Trajectory \uppercase\expandafter{\romannumeral4} marks the conclusion of the cliff flight, with the robot seamlessly transitioning back into wheel mode upon landing. 

These complex missions highlight CapsuleBot's adaptability and proficiency in traversing a variety of terrains and operational modes, demonstrating its adaptive capacity in dynamic and demanding environments.

Some supplementary experiments involved flying with and without the wheels equipped were conducted to verify the effect of the wheel on flight control performance of bicopter. As shown in Fig. \ref{fig:manual_flight}, the tracking of pitch angular velocity when flying with wheels will be negatively affected to some extent. Because the wheel and rotor rotating at the same time increase the inertia of the rotating parts. But the tracking of pitch angular velocity was not significantly affected.

\subsubsection{TRAVERSE ROUGH TERRAINS}
\begin{figure}[t]
	\begin{center}
		\includegraphics[width=1.0\columnwidth]{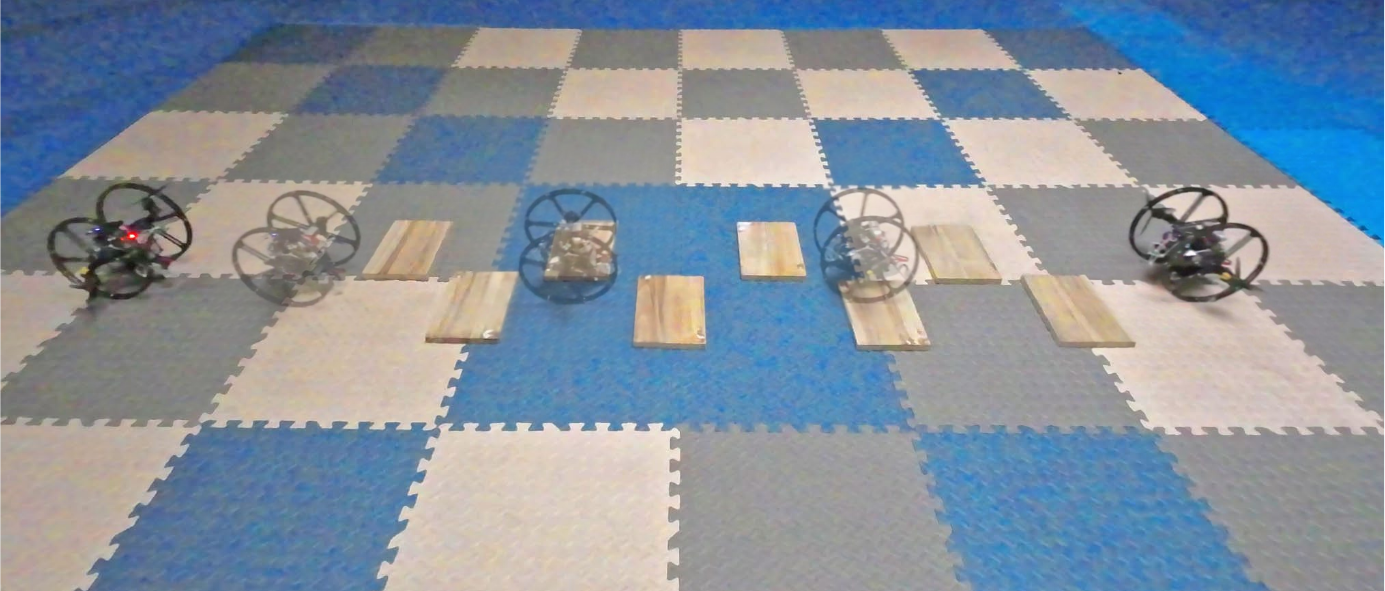}
	\end{center}
	\caption{
		\label{fig:second_task}
		Real-world experiment of traversing rough terrains.
	}
\end{figure}

\begin{figure}[t]
	\begin{center}
		\includegraphics[width=1.0\columnwidth]{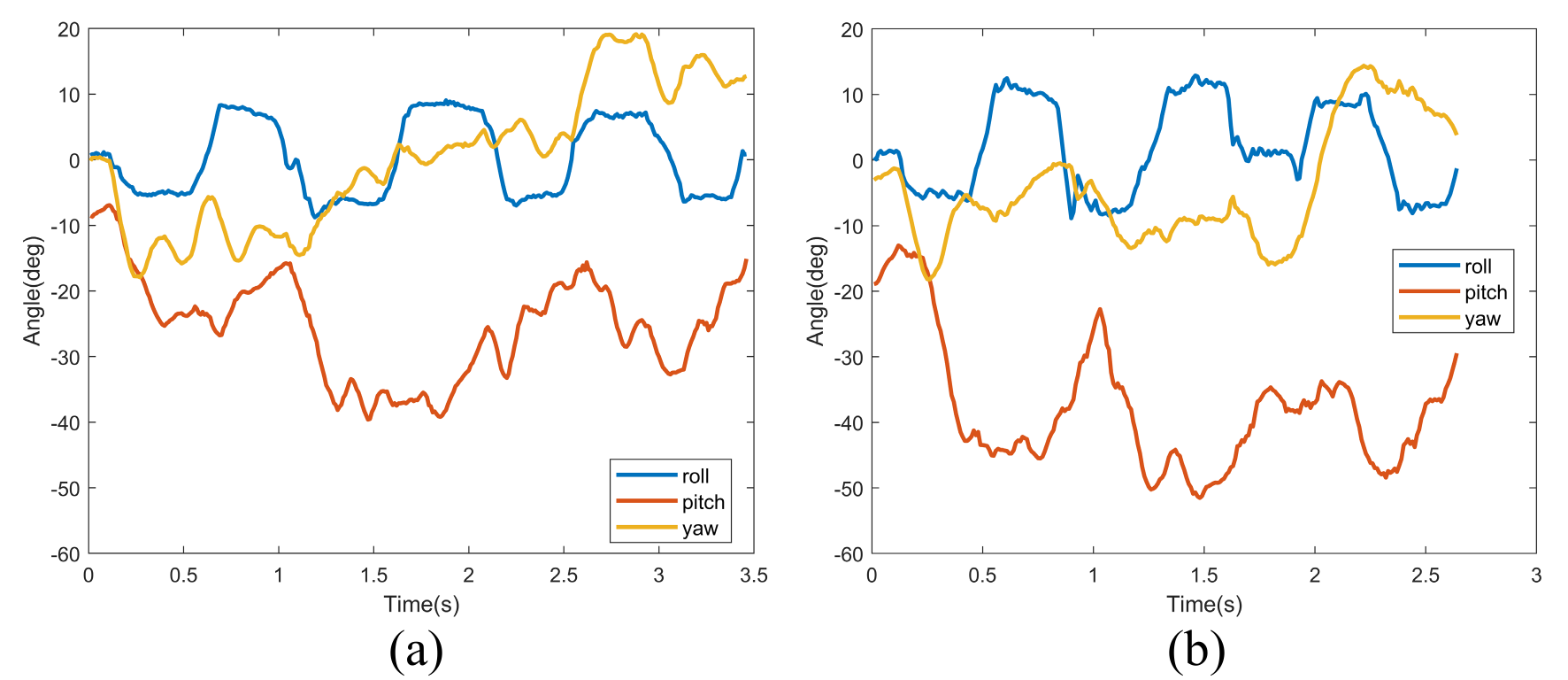}
	\end{center}
	\caption{
		\label{fig:impact_of_traversing_obstacles}
		Attitude of CapsuleBot with or without rotor module in traversing obstacles. (a) Without equipping the blade rotor and brushless motor. (b) Equipping the blade rotor and brushless motor.
	}
\end{figure}

\begin{figure}[t]
	\begin{center}
		\includegraphics[width=1.0\columnwidth]{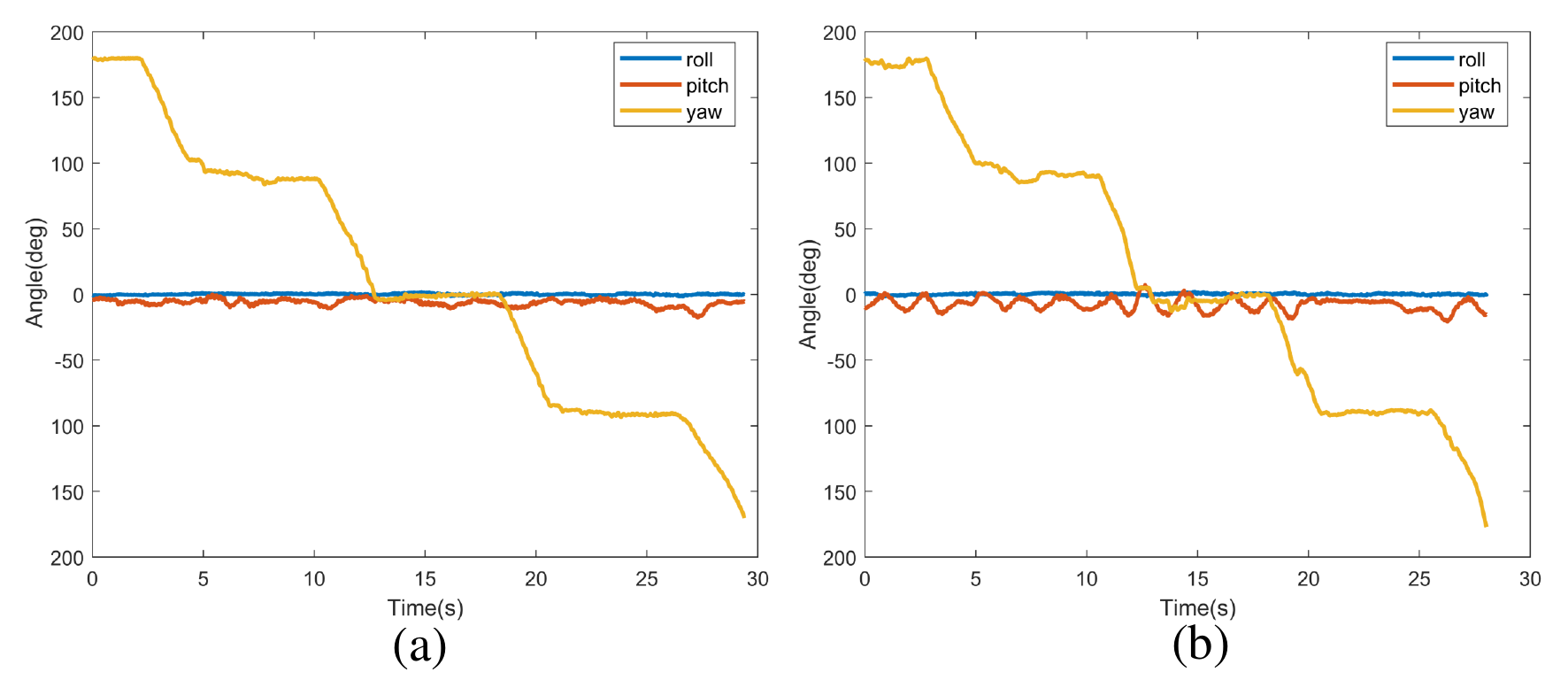}
	\end{center}
	\caption{
		\label{fig:impact_of_ground_mode}
		Attitude of CapsuleBot with or without rotor module in traversing flat ground in a rectangular trajectory. (a) Without equipping the blade rotor and brushless motor. (b) Equipping the blade rotor and brushless motor.
	}
\end{figure}

In the second task scenario (Fig. \ref{fig:second_task}), we simulate a ground surface strewn with wooden block obstacles, each approximately \( 2 \, \mathrm{cm} \) in height, and the maximum step height of wooden block obstacles is approximately \( 3 \, \mathrm{cm} \) when two wheels pass simultaneously. CapsuleBot traverses this terrain steadily in wheel mode. Throughout the experiment, the robot demonstrates remarkable stability and control, maneuvering over the obstacles with precision while maintaining balance and traction. This performance underscores CapsuleBot's ability to traverse rough and unpredictable surfaces, affirming its suitability for deployment in demanding field environments. Some supplementary experiments involved ground mode with and without the rotor module equipped were conducted to verify the effect of the rotor module on movement performance of a two-wheel self-balancing robot. As shown in Fig. \ref{fig:impact_of_traversing_obstacles}, we conducted additional real-world experiments on traversing obstacles, the results show that equipping the rotor module exhibited disadvantages in minimum pitch and range of pitch, as same as traversing flat ground in Fig. \ref{fig:impact_of_ground_mode}. These results validate the rotation of the rotor module does indeed affect the center of gravity of the robot, which in turn impacts pitch stability during wheel mode. However, the impact is limited. We plan to optimize our control strategy to further minimize this effect.

\subsection{Benchmark Comparison}
In this subsection, we select several typical aerial-ground bi-copter robots with different ground mode propulsion systems for comparison, as shown in Table. \ref{tab:benchmark}, representing various design strategies: adaptive structure (Shi's) \cite{shi2024mtabot}, double active wheel-based (Cao's) \cite{cao2023doublebee}, double passive wheel-based with bi-copter transversely arranged (Yang's) \cite{yang2022sytab}, single passive wheel-based (Qin's) \cite{qin2020hybrid}, and double passive wheel-based with bi-copter longitudinally arranged (Lin's) \cite{lin2024skater}. All these robots utilize a bi-copter-based flight mode. We conduct a comparative analysis based on the respective papers, evaluating the following performance indicators.

\begin{enumerate}[]
	\item Number of Motors: CapsuleBot is the only aerial-ground bi-copter robot that achieves both aerial and ground movement using just four motors with active wheels. It integrates an actuated-wheel-rotor design without needing additional motors. In comparison, other configurations with active wheels typically require six motors.
	\item Energy efficiency: Aerial-ground robots with an active wheel configuration on the ground consistently demonstrate superior energy efficiency, with energy consumption rates exceeding 98.9\%. CapsuleBot's favorable energy performance compared to other robots with active wheels can be attributed to its design, which does not require additional motors.
	\item Driving Ground Effect: This refers to whether the ground mode is influenced by the Wing-In-Ground effect, which typically affects robots using rotor propulsion in ground mode, resulting in significant noise and dust, and impacting the control system.
	\item Decoupled Control: This refers to the separate modeling and control of air and ground modes, which may enhance control performance and reduce complexity.
\end{enumerate}

Overall, CapsuleBot offers advantages in low energy consumption, low noise, minimal dust generation, and reduced Wing-In-Ground effect during ground operation. These features highlight its suitability for use in military or field environments as a covert and persistent ISR platform.

\section{CONCLUSION}\label{sec:conclu}
CapsuleBot is a novel hybrid aerial-ground bi-copter robot designed for long-endurance, low-noise operations. It combines the agility of a bi-copter in flight with the efficiency of a two-wheel self-balancing robot on the ground, utilizing an innovative actuated-wheel-rotor design. This unique design enables hybrid propulsion using only four motors, eliminating the need for additional motors found in conventional bi-copters. The study includes the development of dynamics and control systems for both aerial and ground modes, validated through experiments that demonstrate low power consumption and minimal noise emissions during ground operations. CapsuleBot excels in challenging terrains, such as climbing steep inclines, flying over cliffs, and traversing rugged landscapes, showcasing its suitability for implementation in military or field environments as a covert and persistent ISR platform. In our future work, we will focus on developing an autonomous multimodal path planning method.

\begin{table*}[t]
	\renewcommand\arraystretch{1.2}
	\setlength{\tabcolsep}{3pt}
	\centering
	\caption{Performance comparison with several air-ground robots that utilize a bi-copter-based flight mode.}
	\label{tab:benchmark}
	\vspace{-0.2cm}
	\includegraphics[width=2.0\columnwidth]{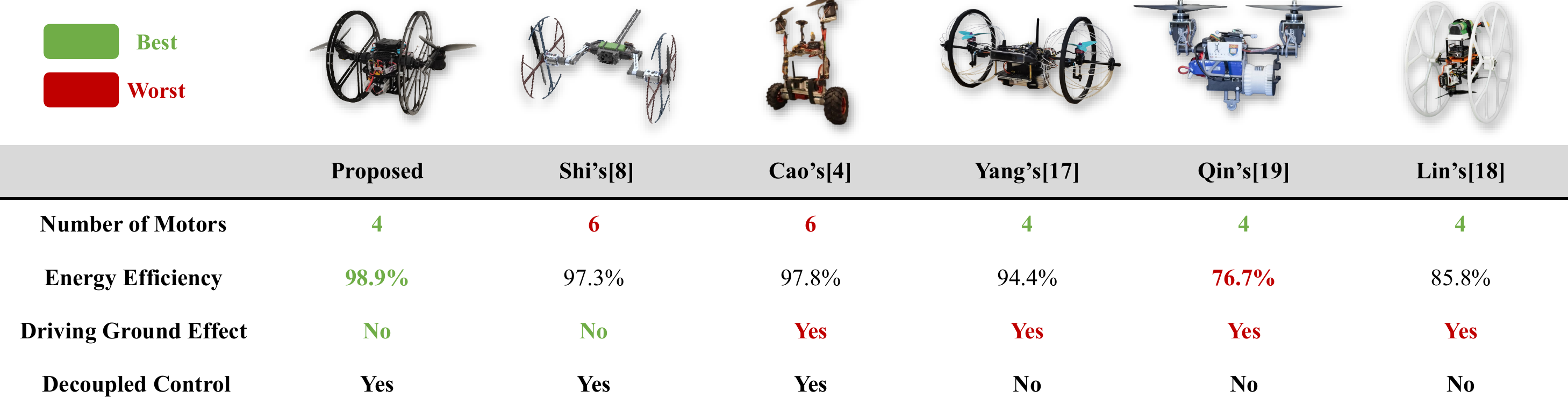}
\end{table*}

\bibliography{24-2186-final}

\end{document}